\documentclass[conference]{IEEEtran}
\IEEEoverridecommandlockouts
\usepackage{cite}
\usepackage[cmyk]{xcolor}
\usepackage{amsmath,amssymb,amsfonts}
\usepackage{algorithmic}
\usepackage{graphicx}
\usepackage{textcomp}
\usepackage{CJKutf8}
\usepackage{float}
\usepackage{color,soul}
\usepackage{times}
\usepackage{pifont}
\usepackage{bbding}

\def\BibTeX{{\rm B\kern-.05em{\sc i\kern-.025em b}\kern-.08em
    T\kern-.1667em\lower.7ex\hbox{E}\kern-.125emX}}

\begin{document}





\title{QLSC: A Query Latent Semantic Calibrator for Robust Extractive Question Answering\\
\thanks{${\dagger}$ \;Equal contribution.}
\thanks{{\Envelope} Corresponding author: Xulong Zhang (zhangxulong@ieee.org).}
}

\author{\IEEEauthorblockN{Sheng Ouyang$^{\dagger}$, Jianzong Wang$^{\dagger}$, Yong Zhang, Zhitao Li, Ziqi Liang, Xulong Zhang\textsuperscript{\Envelope}, Ning Cheng, Jing Xiao}
\IEEEauthorblockA{\textit{Ping An Technology (Shenzhen) Co., Ltd., China} }
}

 \maketitle


\begin{abstract}
Extractive Question Answering (EQA) in Machine Reading Comprehension (MRC) often faces the challenge of dealing with semantically identical but format-variant inputs. Our work introduces a novel approach, called the ``Query Latent Semantic Calibrator (QLSC)'', designed as an auxiliary module for existing MRC models. We propose a unique scaling strategy to capture latent semantic center features of queries. These features are then seamlessly integrated into traditional query and passage embeddings using an attention mechanism. By deepening the comprehension of the semantic queries-passage relationship, our approach diminishes sensitivity to variations in text format and boosts the model's capability in pinpointing accurate answers. Experimental results on robust Question-Answer datasets confirm that our approach effectively handles format-variant but semantically identical queries, highlighting the effectiveness and adaptability of our proposed method.
\end{abstract}


\section{Introduction}

    


Machine Reading Comprehension tasks, specifically Extractive Question Answering, are essential for natural language understanding and have gained significant attention in recent years\cite{guan2022block, hao2022recent}. EQA aims to develop models that can accurately answer questions by extracting relevant information from a given passage. In contrast to generative models\cite{Zeng2022GLM130BAO}, extractive models are more advantageous in finding the most relevant answers within a defined text than in generating answers. This distinction serves to mitigate the likelihood that the model generates hallucinatory responses, particularly in resource-constrained environments. However, EQA models also face the challenge of handling semantically identical but format-variant input. The format-variant question, also called a paraphrase question, retains the same meaning as the original question but varies in its phrasing or wording. This challenge arises when questions with the same underlying meaning are expressed using different words or syntactic structures.


Attempts have also been made to enhancing the robustness of language models within the realm of EQA. Adversarial attacks and adversarial training methods have been explored to enhance the model's capacity in addressing format variations. Adversarial attacks involve the generation of perturbations or modifications of the input data to mislead the predictions of the model\cite{dong2023adversarial, le2022global, Bartolo2021ImprovingQA}. These attacks can reveal vulnerabilities in the model and highlight the need for improved robustness \cite{Li2023AnnealingGP, Alzantot2018GeneratingNL, talmor2019multiqa}. Adversarial training\cite{ziegler2022adversarial, pan2022improved, zhou2022enhancing}, on the other hand, involves training the model in both clean and adversarial examples, forcing it to become more resilient to perturbations. However, these methods often require substantial amounts of data, which can limit their practicality.

In this study, we introduce an innovative method referred to as the ``Query Latent Semantic Calibrator'' to address the challenge of format-variant input in EQA. The method focuses on learning and imparting an embedding that enhances the model's robustness. Inspired by Lin et al.\cite{lin2018nextvlad}, the approach comprises Semantic Center Learning, Soft Semantic Feature Selection, and Query Semantic Calibration. In Semantic Center Learning, A set of randomly initialized information features is used to learn various subspace semantic center features of all queries. Soft Semantic Feature Selection fuse subspace semantic center features to get the global latent semantic center features as the semantic embedding. Similar to soft prompts, which are then integrated into the vanilla queries and passage embedding using an attention mechanism to reduce its sensitivity to variations in text format, helping the model better understand the association between the query and the passage, enabling more accurate answer extraction. This plug-in method, integrated with various encoders, improves the robustness within the traditional extractive framework. Unlike other methods that require additional training data, knowledge, or model complexity, we achieve effectiveness by simply adding auxiliary modules, emphasizing the efficacy and adaptability of the Query Latent Semantic Calibrator.


In summary, our contributions are as follows: 
\begin{itemize}
\item We propose the ``Query Latent Semantic Calibrator'', a novel auxiliary module that enhances EQA model robustness. This module is specifically designed to integrate latent semantic center features into traditional queries and passage embeddings.

\item Our approach innovatively employs a scaling strategy and soft select strategy to address format-variant challenges in EQA. Through these strategies, multiple potential semantic features can be generated in different subspaces and combined to obtain the final latent semantic center feature.

\item Extensive experiments were carried out on robust QA datasets, and the findings demonstrate the superior performance of EQA models that are equipped with the Query Latent Semantic Calibrator. Comparative analyses additionally emphasize the improved robustness of our approach.
\end{itemize}

\begin{figure}[t]
  \begin{center}
  \includegraphics[width=\linewidth]{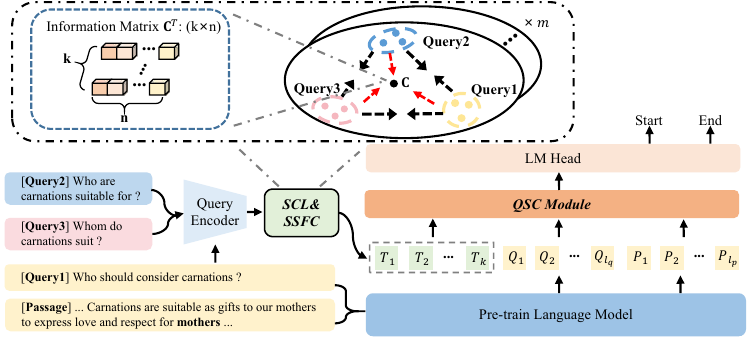}
  \end{center}
    \caption{The architecture of our proposed QLSC method. SCL represents Semantic Center Learning, SSFS represents Soft Semantic Feature Selection, and QSC represents Query Semantic Calibration. $C$ is the information matrix and $m$ is the number of subspaces. $k$ is the number of the semantic center feature $T$. $l_{q}$ and $l_{p}$ respectively represent the length of the query and the passage encoded by the pre-trained language model.}
  \label{fig:pic1}
\end{figure}

\section{Related Work}
\subsection{Extractive Question Answering}
The machine reading comprehension task requires the design of a model to answer the set question with the given textual context information\cite{kazi2023survey}. This objective can be viewed as tackling two separate multiclass classification tasks, aiming to predict the starting and ending positions of answer spans. Its main tasks are divided mainly into the following four types\cite{baradaran2022survey,liu2019neural}. The first is the Cloze task. The second is a multiple choice task. The third is the segment extraction task, which requires a given piece of text and a question and requires the model to extract a continuous subsequence from the text as an answer. The fourth is the free answering task. With the continuous progress of deep learning, the development of the MRC dataset\cite{zeng2020survey,tan2021gcrc,han2021ester} and the attention mechanism\cite{bahdanau2014neural} have greatly promoted the improvement of the level of various tasks in MRC. 


In the past few years, a growing tendency has emerged to convert NLP tasks into extractive question answering formats\cite{Namazifar2020LanguageMI}. McCann et al.\cite{mccann2018natural} transformed NLP tasks such as summarization or sentiment analysis into question answering. Li et al.\cite{li2020unified} uses different natural language questions to describe each type of entity, and entities are extracted by answering these questions according to the contexts. For example, the question `which person is mentioned in the text' is used for the PER(PERSON) label. Li et al.\cite{li2020event} introduced a novel paradigm for the entity-relation extraction task, which defines it as a multi-round question answering task. Each entity and relationship type correspond to a question answer template, each entity and corresponding relationship are extracted using question and answer tasks.


\begin{table}[ttb]
\begin{CJK*}{UTF8}{gbsn}
  \caption{An example illustrates the over-sensitivity question.}
  \centering
  \setlength{\tabcolsep}{1.5mm}
  \begin{tabular}{|l|}
    \hline
    \textbf{Context:}\\ ... 玫瑰适合送情侣，可能可以创造美好的爱情故事。康乃馨适合作\;\;\;\;\\为送给母亲的礼物，表达对母亲的爱和尊重。...
    \\
    \;... Roses are suitable for gifting lovers and may create beautiful love\\ storys. Carnations are suitable as gifts to our mothers to express love\\ and respect for mothers ...\;
    \\ \hline
    \textbf{Original Question}: 康乃馨送给什么人合适? \\(Who are carnations suitable for?)\\ \textbf{Golden Answer}: 母亲\;(mothers)\\ \textbf{Predicted Answer} : \;RoBERTa: 母亲\;(mothers) \;Ours:\;母亲\;(mothers)                          \\  \hline
    \textbf{Paraphrase Question:}\;康乃馨可以送给谁? \\(Who can you give carnations to?)\\ \textbf{Golden Answer}: 母亲\;(mothers)\\ \textbf{Predicted Answer}: \;RoBERTa: 情侣\;(lovers) \;Ours:\;母亲\;(mothers)\\              
    \hline
  \end{tabular}  
  \label{ppl:sensitivity}
\end{CJK*}
\end{table}

\begin{table}[ttb]
\begin{CJK*}{UTF8}{gbsn}
  \caption{An example illustrates the over-stability question.}
  \centering
  \setlength{\tabcolsep}{1.1mm}
  \begin{tabular}{|l|}
    \hline
    \textbf{Context:}\\ ... 大多数的宝宝，白天基本要睡2至3次，一般是上午睡1次，下午睡\\1至2次，每次1至2小时不等。夜间一般要睡10小时左右。...
    \\ 
    ... Most babies sleep 2 to 3 times during the day, usually 1 time in the \\ morning and 1 or 2 times in the afternoon, ranging from 1 to 2 hours \\each time. Usually, sleep about 10 hours at night ... 
    \\ \hline
    \textbf{Original Question}: 大多数宝宝白天上午要睡几次? \\(How many times do most babies sleep in the morning during the day?)\\ \textbf{Golden Answer}: 1次\;(1 time)\\ \textbf{Predicted Answer} : \;RoBERTa: 1次\;(1 time) \;Ours:\;1次\;(1 time)                          \\  \hline
    \textbf{Paraphrase Question:}\;大多数宝宝白天下午要睡几次? \\(How many times do most babies sleep in the afternoon during the day?)\\ \textbf{Golden Answer}: 1至2次\;(1 or 2 times)\\ \textbf{Predicted Answer}: \;RoBERTa: 1次\;(1 time) \;Ours:\;1至2次\;(1 or 2 times)\\              \hline
  \end{tabular}  
  \label{ppl:stability}
\end{CJK*}
\end{table}

\subsection{EQA System Robustness}
In this work, our work studied three aspects of the Dureader$_{robust}$ dataset introduced by Tang et al.\cite{tang2021dureader_robust}. More fine-grained metrics are used to quantify the robustness of the EQA system. Specifically, the over-sensitivity aspect, over-stability aspect, and generalization aspect. It is worth mentioning that the instances in Dureader$_{robust}$ are real natural Chinese text, not modified unnatural text. 
As shown in Table~\ref{ppl:sensitivity}, The original question and the paraphrase question have the same meaning but differ in the sample description. The findings indicate that RoBERTa accurately answers the original question, but provides an incorrect response to the rephrased question. In contrast, our approach yields accurate answers for both questions. The over-sensitivity aspect focuses on evaluating models' perturbation of outputs when handling paraphrased questions. As indicated in Table~\ref{ppl:stability}, the original question and the paraphrase question have different meanings but are similar in the description of the sample. The results show that the same answers are obtained using only the RoBERTa model, but the answer is wrong for the paraphrased problem. The over-stability aspect investigates the model's ability to discern the difference in question utterance that leads to different outputs. The generalization aspect focuses on measuring the models' ability to answer out-of-domain questions.

Adversarial training is a technique that can significantly enhance the robustness and generalization capability of MRC models. FGSM\cite{goodfellow2014explaining} introduces small perturbations to the input. Increase the disturbance in the direction of increasing the loss. Specifically, gradient ascent is performed on the input embedding layer.
After that, FGM\cite{miyato2016adversarial} further optimizes the FGSM method and thus obtains better adversarial inputs. Madry et al.\cite{madry2018towards} summarize previous work and unify the adversarial training format. Furthermore, they identify a reliable way to train and attack neural networks. Although previous adversarial training methods achieve great results, they consume more computing resources. Therefore, FreeAT\cite{shafahi2019adversarial} was proposed, which optimizes the training speed based on PGD. In addition, many related works\cite{jiang2020smart, zhang2019you, zhu2019freelb} also optimize adversarial training effectively.

\vspace{1mm}
\section{Methodology}
\vspace{2mm}

\label{sec:format}
The entire QLSC framework can be depicted in Figure~\ref{fig:pic1}. Our method comprises three main stages: Semantic Center Learning (SCL), Soft Semantic Feature Selection (SSFS), and Query Semantic Calibration (QSC). Semantic Center Learning involves ingesting original information features and queries and generating multiple latent semantic center features across different subspaces. Soft Semantic Feature Selection adjusts these subspace semantic center features by giving higher weights to more semantically robust ones, while downplaying less informative ones, and then fuse them to get a global latent semantic center feature. Query Semantic Calibration uses the attention network fuses all the features to achieve query calibration.

\begin{table}[b!]
  \label{propose_algorithm}
  \centering
  \setlength{\tabcolsep}{0.3mm}
  \small
  \begin{tabular}{l}
    \hline
    \textbf{Algorithm 1} The Proposed QLSC Algorithm \\
    \hline
    \textbf{Require}:
    Query sentence $Q$, information matrix $C$ \\
    \textbf{1}: Randomly initialized $C$ \\ 
    \textbf{2}: Encoding the $Q$ to query feature matrix $H$ \\

    \textbf{3}: Using a linear network $W$ to calculate $H^{'}$ and $C^{'}$\\

    \textbf{4}: Dimensional expansion and transformation to obtain $\tilde{H}$ and $\tilde{C}$\\

    \textbf{5}: $k$ = 1, $i$ = 1\\
    \textbf{6}: \textbf{for} $k$ = 1, ... , k \\
    \textbf{7}: \;\;\;\;\textbf{for} $i$ = 1, ... , m \\
    \textbf{8}: \;\;\;\;\;\;\;\;Set $w_{ik}$ and $b_{ik}$ as training parameters\\
    \textbf{9}: \;\;\;\;\;\;\;\;$f_{ik}=softmax(w_{ik}^T\tilde{H}+b_{ik})$\\

    \textbf{10}: \;\;\;\;\;\;\;\;$s_{k}^i=f_{ik}(\tilde{H}_i-{\tilde{C}_{k}})$  	\\

    \textbf{11}: \;\;\;\;\;\;\;\;$v_{k}^i=sigmoid(\tilde{H}_i)\cdot s_{k}^i$\\

    \textbf{12}: \;\;\;\;\;\;$T_{k}=T_{k} + v_{k}^i$ \\

    \textbf{13}: \;\;\;\;\;\;$i = i + 1$\\

    \textbf{14}: \;\;\;$k = k + 1$\\

    \textbf{15}: $T=\{T_{1}, T_{2},\cdots, T_{k}\}$\\

    \textbf{16}: Using the attention mechanism $score$ (Equation (7)) to output \\ \;\;\;\;\;\; the query semantic calibration features $Q$ (Equation (8))\\

    \hline
    
 \end{tabular}  
\end{table}

\subsection{Semantic Center Learning}
\vspace{1mm}
Semantic Center Learning incorporates two primary components: the Subspace Mapping Network and the Query Semantic Information Fusion. Information vectors comprise high-dimensional features that encapsulate various semantic attributes of queries. In this learning approach, both query and information vectors are mapped to distinct subspaces. Subsequently, they are merged to establish potential semantic centers within these subspaces. Through this process, the resulting latent semantic center features can represent queries with the same meaning but in varied formats, capturing a diverse range of high-level query information. 

\subsubsection{Subspace Mapping Network}
Subspace Mapping Network expands the richness of features by mapping query and information vectors into multiple subspaces, which is beneficial for mining semantic feature centers from multiple different perspectives.
Given a query sentence $\mathbf{\mathcal{Q}}$, input it into the LSTM to obtain its encoded feature matrix: $H=\left[h_{1} , \cdots , h_{l}\right]\in R^{n \times l}$, where $l$ is the sequence length, $n$ is the hidden state dimension and $h_{i} \in R^{n}$ is the query vector of each token. $C=\left[c_{1} , \cdots , c_{k}\right] \in R^{n \times k}$ is a matrix carried out by information vectors $c_{i}$, where $c_{i}\in R^{n}$ and $k$ is the number of information vectors. $C$ is randomly initialized. The Subspace Mapping Network maps the initialized information vectors $\left\{c_{i}\right\}_{i=1}^{k}$ and the encoded query vectors $\left\{h_{i}\right\}_{i=1}^{l}$ to $m$ different subspaces: $G=\left\{g_{1}, ..., g_{m}\right\}$, creating diverse and rich feature representations.

Specifically, a linear scaling network represented by $W\in{R}^{mn\times n}$ is used to increase the hidden dimension of $H$ and $C$. The transformation $H^{'} = W H$ converts $H\in{R}^{n \times l}$ to $H^{'}\in{R}^{mn\times l}$, ${C}^{'} = W C$ converts $C\in{R}^{n \times k}$ to ${C}^{'}\in{R}^{mn\times k}$. $H^{'}$ represents the transfored query matrix, and ${C}^{'}$ represents the transfored information matrix. Subsequently, the extended matrices are segmented in the direction of the column and dimensionally transformed to create a new query matrix $ \tilde{H} \in{R}^{m\times n\times l}$, where $\tilde{H}=[\tilde{H}_{1}, \cdots ,\tilde{H}_{l}]$ and $ \tilde{H}_{i} \in R^{m \times n}$. Continue the same operation to get a new information matrix $\tilde{C} = \left [\tilde {C_{1}}, ..., \tilde {C_{k}}\right ] \in{R}^{m\times n\times k}$.

\subsubsection{Query Semantic Information Fusion}
Query Semantic Information Fusion generates rich subspace Semantic center features by integrating the query embedding feature with the initialized information feature. This process ensures that the semantic diversity of the input queries and latent semantic center are reflected in the subspace semantic center features.

Specifically, in the $i$th group, the soft attention mechanism corresponding to the $k$ information vector is computed as follows.

\begin{equation}
  f_{ik}\left(\tilde{H}_{i}\ \right)=\frac{e^{w_{ik}^T\tilde{H}+b_{ik}}}{\sum_{k=1}^{K}e^{w_{ik}^T\tilde{H}+b_{ik}}} 	   	 		      		 
\end{equation}
%
where $w_{ik}$ and $b_{ik}$ are trainable parameters.

The subspace center features $s_{k}^i$ is calculated from the difference between $\tilde{H}_i$ and $\tilde{C}_{k}$ by formula $\left (1\right )$ and as follows:

\begin{equation}
  s_{k}^i=f_{ik}(\tilde{H}_i-{\tilde{C}_{k}})	   	  
\end{equation}

\vspace{1mm}
\subsection{Soft Semantic Feature Selection}
\vspace{1mm}
Soft Semantic Feature Selection uses an attention mechanism to selectively assimilate the most informative semantic features of queries. This process results in a robust latent semantic center feature. Such a latent embedding adeptly captures the intricate and high-dimensional semantics of the query, mitigating the noise introduced by input perturbations.

The calibration mechanism is computed as follows:
\begin{equation}
\alpha_{i}=sigmoid(w_i^T\tilde{H}_{i}+b_i)
\end{equation}
where $w_i^T$ and $b_i$ are trainable parameters.

The subspace-adjusted semantic center feature $v_{k}^i$ is calculated as the product of $\alpha_i$ and the subspace center feature vector $s_{k}^i$.

\begin{equation}
v_{k}^i=\alpha_{i} s_{k}^i
\end{equation}

Finally, the global latent semantic center feature $T$ is computed by summing all $v_{k}^i$.

\begin{equation}
T_{k}=\sum_{i} v_{k}^i
\end{equation}

\vspace{-2mm}

\begin{equation}
T=\{T_{1}, T_{2},\cdots, T_{k}\} \in {R}^{k\times n}
\end{equation}


\subsection{Query Semantic Calibration}
The attention mechanism is used to merge the global latent semantic center feature $T$ and the query feature $Q$ to achieve query calibration. This integration process enhances the model's ability to comprehend and respond to semantically equivalent queries despite their format variations.

The dot product is computed and softmax is applied to obtain the scores $score_{r,j}$ for $T_j$ and $Q_r$.
\vspace{1mm}
\begin{equation}
  score_{r,j}=\frac{e^{{Q_r}{T_j}}}{\sum_{j}e^{{Q_r}{T_j}}}
\end{equation}

\vspace{1mm}
Finally, the features of the attention calculation are integrated into the features of the token through a sum operation.

\begin{equation}
  Q_r= Q_r +\sum_{j}score_{r,j}{T_j}
  \label{eqh}
\end{equation}

Similarly, the same attention process is used for the global latent semantic center feature $T$ and the passage feature $P$ to enhance $T$'s understanding of textual information.

\vspace{-1mm}
\subsection{Loss Function}
The starting and ending positions of each token are obtained through two full connection layers (LM Head), respectively. Therefore, the cross-entropy loss function is used. Add the starting position loss function and the ending position loss function as the total loss.

\section{Experiment and Analysis}
\label{sec:experiment}

\begin{table*}[htbp]
  \caption{
  The results of all compared models on the development set and three test sets under the Dureader$_{robust}$.}
  \centering
  \setlength{\tabcolsep}{2.5mm}
  \begin{tabular}{lccccccccc}
    \hline
      &&\multicolumn{2}{c}{Dev set} &\multicolumn{2}{c}{Over Sensitivity} &\multicolumn{2}{c}{Over Stability} &\multicolumn{2}{c}{Generalization}  \\
     Model &+QLSC & F1 & EM &F1 & EM & F1 & EM & F1 & $EM$\\ 
    
    \hline
    {BERT} &\ding{56} &77.68 &64.64 &54.03 &37.92 &50.93 &30.82 &58.54 &43.34 \\
    {BERT} &\ding{52} &${80.43}_{\textcolor{green}{+2.75}}$ &${67.89}_{\textcolor{green}{+3.25}}$ &$55.61_{\textcolor{green}{+1.58}}$ &$40.59_{\textcolor{green}{+2.67}}$ &$56.70_{\textcolor{green}{+2.75}}$ &$36.49_{\textcolor{green}{+5.67}}$ &$61.44_{\textcolor{green}{+2.9}}$ &$45.17_{\textcolor{green}{+1.83}}$ \\
    \hline
    {RoBERTa} &\ding{56} &84.22 &71.91 &60.74 &43.80 &58.05 &37.70 &64.20 &46.24\\

    {RoBERTa} &\ding{52} &${85.49}_{\textcolor{green}{+1.27}}$ &${73.89}_{\textcolor{green}{+1.98}}$ &$61.37_{\textcolor{green}{+0.63}}$ &$44.69_{\textcolor{green}{+0.89}}$ &$65.28_{\textcolor{green}{+7.75}}$ &$46.77_{\textcolor{green}{+9.07}}$ &$66.20_{\textcolor{green}{+2.00}}$ &$48.84_{\textcolor{green}{+2.60}}$ \\
    \hline
    {MacBERT} &\ding{56} &85.54 &73.39 &60.01 &41.47 &61.07 &41.53 &66.27 &47.88\\
    {MacBERT} &\ding{52} &${85.75}_{\textcolor{green}{+0.21}}$ &${74.81}_{\textcolor{green}{+1.42}}$ &$61.26_{\textcolor{green}{+1.25}}$ &$44.25_{\textcolor{green}{+2.78}}$ &$64.96_{\textcolor{green}{+3.89}}$ &$48.39_{\textcolor{green}{+6.86}}$ &$66.69_{\textcolor{green}{+0.42}}$ &$50.19_{\textcolor{green}{+2.31}}$ \\
    \hline
    {PERT} &\ding{56} &86.24 &75.16 &63.59 &46.13 &66.72 &47.78 &67.75 &49.81\\
    {PERT} &\ding{52} &${87.48}_{\textcolor{green}{+1.24}}$ &${76.23}_{\textcolor{green}{+1.07}}$ &$65.65_{\textcolor{green}{+2.06}}$ &$47.84_{\textcolor{green}{+1.71}}$ &$71.84_{\textcolor{green}{+5.12}}$ &$55.04_{\textcolor{green}{+7.26}}$ &$69.19_{\textcolor{green}{+1.44}}$ &$50.19_{\textcolor{green}{+0.77}}$ \\
    \hline

    \hline
  \end{tabular}  
  \label{tab:table1}

\end{table*}

    

\vspace{1mm}
\subsection{Datasets}
\vspace{1mm}
Experiments are conducted on the basis of the reconstructed Dureader$_{robust}$ dataset and the SQuAD1.1 dataset.

\textbf{Dureader$_{robust}$}: Our experiments are carried out using the Dureader$_{robust}$ dataset, which includes training set, development set, and test set with sizes of 15k, 1.4k, and 1.3k. It also introduces three challenge subsets, called over-sensitivity, over-stability, and generalization, to evaluate robustness and generalization of the system. The data set comprises questions and passages as input and answers as labels. As Tang et al.\cite{tang2021dureader_robust} do not provide annotated answers to the challenge sets, the performance of different models is validated based on the reconstructed subsets introduced by Li et al.\cite{li2021robustness}. 
The over-sensitivity subset consists of multiple paraphrased questions with different expressions with the same semantics for the same answer. The over-stability subset consists of samples with a large number of interfering sentences. That is, multiple sentences in the same sample have multiple identical words. The generalization subset constructs other domain data in addition to the in-domain data. 

\textbf{SQuAD1.1}: Our method is evaluated on the SQUAD1.1 dataset. The passages in the SQuAD1.1 dataset are obtained through retrieval from Wikipedia articles. Each question within the data set is designed to generate an answer that can be directly extracted as a contiguous span from the provided passage. The data set is divided into training, development and test sets, with approximate sizes of 88k samples for training, 11k for development and 10k for the test set.

\subsection{Experimental Setups}
\vspace{1mm}
In the Dureader$_{robust}$ dataset experiment, we employed chinese BERT$_{base}$\cite{cui2021pre}, chinese RoBERTa$_{large}$\cite{cui2021pre}, chinese MacBERT$_{large}$\cite{cui2020revisiting} and PERT$_{large}$\cite{cui2022pert} with our QLSC module. The batch size is configured as 4, and Adam optimizer is applied with a learning rate of 3e-5. The input question and passage are restricted to a maximum length of 64 and 512, respectively, the maximum length of the generated answer is 30. Evaluation was carried out using random seed 42. During the conduct of comparative experiments, all parameters are kept consistent. The evaluation is based on the F1 and EM values\cite{rajpurkar2016squad}. In random seed experiments, different seed numbers are used, with each seed's experiment conducted five times. The final results are calculated as averages.

In the SQuAD1.1 dataset experiment, BERT\cite{devlin2019bert} was used as a baseline. Our experiments using the adversarial training method BERT+AT+VAT\cite{yang2019improving}, KT-NET\cite{Yang2019EnhancingPL}, XLNet\cite{Yang2019XLNetGA}, RoBERTa\cite{Yinhan2019RoBERTa} and our QLSC method. All parameter settings must be consistent with those described in the text.

    
   

\subsection{Main Results}
The performance results of different pre-trained lanauage models on the $Dureader_{robust}$ dataset are presented in Table~\ref{tab:table1}. The test set of Over Sensitivity and Over Stability together reflect the robustness of different models, and Generalization reflects the generalization ability on the test set. As shown in this table, with the introduction of our QLSC module, all models achieved a certain improvement on the dev set and three test sets.

The results represent our QLSC method, which can help the EQA model improve its understanding of the query-passage association by integrating latent semantic center features into the original embedding. The results of generalization dataset could support that our QLSC method can be well adapted to out-of-domain data, and has a positive effect on the extraction of context semantics. 



Training and testing were conducted on the SQuAD1.1 data, and the experimental results are presented in Table~\ref{tab:tableaquad}.

\begin{table}[htbp]
  \caption{Different model results on SQuAD test set.}
  \centering
  \setlength{\tabcolsep}{3.1mm}
  \begin{tabular}{lcccccc}
    \hline
     Model && + QLSC && EM && F1 \\ 
     \hline 
    Human Performance && - &&82.3 &&91.2\\
    \hline 
    BERT &&\ding{56} &&85.1 &&91.8\\
    BERT+AT+VAT(2.0) &&\ding{56} &&86.9 &&92.6 \\   
    KT-NET &&\ding{56} &&85.9 &&92.4\\
    XLNet &&\ding{56} &&89.9 &&95.1\\
    RoBERTa &&\ding{56} &&90.4 &&95.3\\
    \hline
    \textit{Our method} \\
    BERT &&\ding{52} &&87.5 &&94.2 \\
    RoBERTa &&\ding{52} &&\textbf{93.1} &&\textbf{96.8}\\
    \hline
  \end{tabular}  
  \label{tab:tableaquad}
  \vspace{-1mm}
\end{table}

The results of the random seed experiments indicate that when the random seed is set to 42 and our QLSC module is used, the F1 and EM values are 85.49 and 73.89, respectively. When only RoBERTa$_{large}$ is used, the F1 and EM values are 84.22 and 71.91, respectively. Furthermore, the F1 and EM values consistently demonstrate strong performance in other random seeds. These experimental findings highlight the robustness of our QLSC method, which performs better under different random seed settings.

\vspace{-1mm}
\subsection{Ablation Study}
\subsubsection{Number of Information Features}

\begin{figure}[t]
  \begin{center}
  \includegraphics[width=\linewidth]{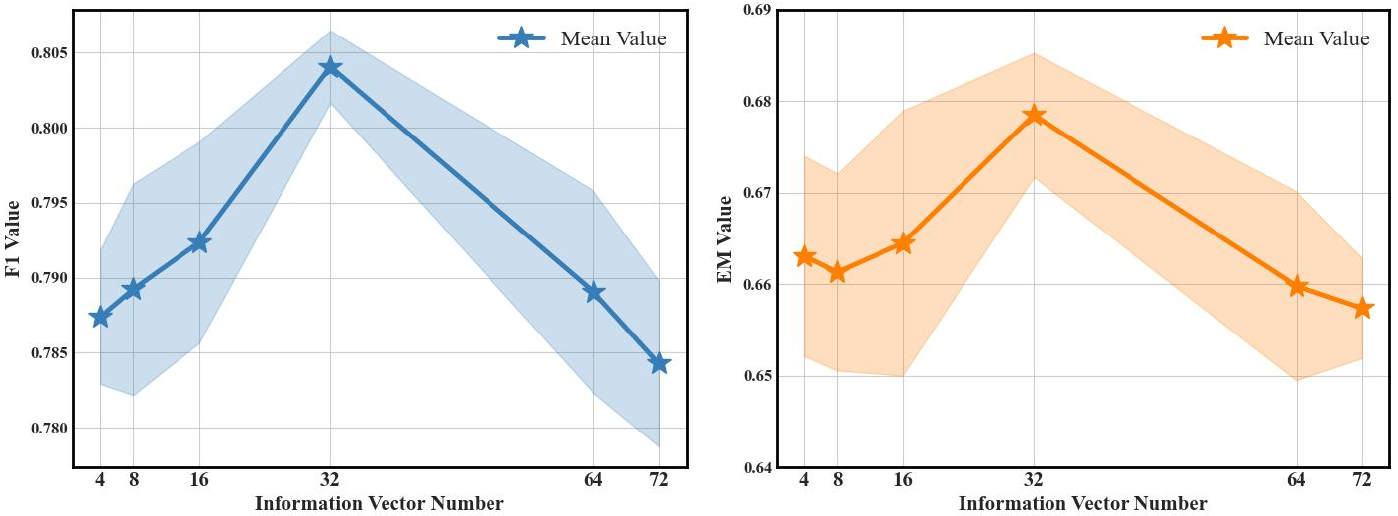}
  \end{center}
    \caption{The effect of a different number of information features on model performance on Dev set.}
  \label{tab:table2}
\end{figure}

On Dureader$_{robust}$, experiments were conducted to observe how varying the number of information features $K$ affected the model performance when using BERT${base}$ backbone. As depicted in Figure~\ref{tab:table2}, with the increase of $K$, the F1 and EM values initially rise before exhibiting a subsequent decline. When the value $K$ was set to 32, the model's F1 and EM values reached the maximum.

If the value of $K$ is too small, the semantic richness of the potential semantic centers calculated from information features is low, and therefore the semantic calibration effect on the query is poor. An excessively large value of $K$ will lead to noise in the information features, thus affecting the potential semantic center's calibration of the query's semantics.

In summary, the appropriate $K$ can help the model to better calibrate queries' utterance perturbations while preserving the ability to discern the nuance in queries expression that may have different queries intentions.

\begin{figure}[t]
  \begin{center}
  \includegraphics[width=\linewidth]{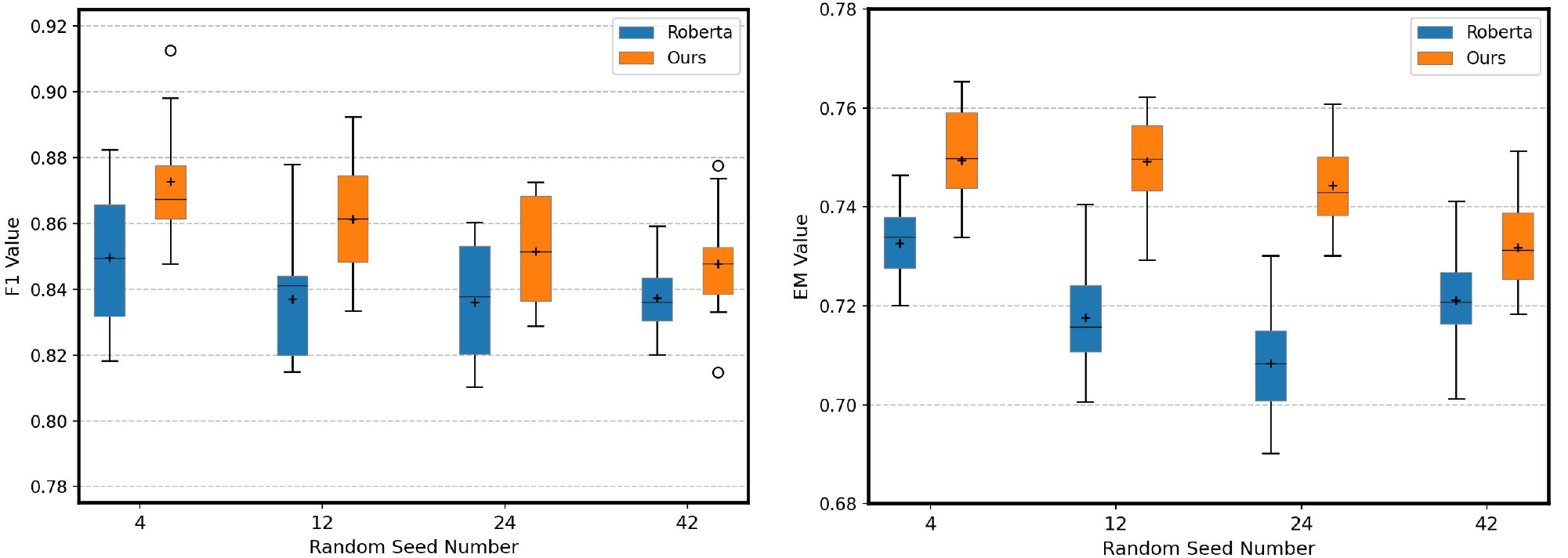}
  \end{center}
  \caption{The influence of various random seeds on model performance.}
  \label{tab:table5}
\end{figure}

    


\subsubsection{Random Seed Experiments}
Displayed in Figure~\ref{tab:table5}, random seed experiments have been carried out. Ten experiments were carried out for each random seed. The experiments demonstrate that, under different random seeds, the use of the QLSC makes the original model more stable and yields better performance.

\begin{table}[htbp] 
  \centering
  \caption{Results on the test sets when using different query encoders.
  }
  \setlength{\tabcolsep}{4.1mm}{
  \begin{tabular}{lcccccc}
    \hline
    &\multicolumn{2}{c}{Over Sensitivity} &\multicolumn{2}{c}{Over Stability} \\
     Question Encoder & $F1$ & $EM$ & $F1$ & $EM$ \\ 
    
    \hline
    CNN$_{kelnel=5}$ &58.57 &42.55 &63.69 &45.36 \\
    CNN$_{kelnel=7}$ &62.00 &43.66 &59.49 &40.73 \\
    LSTM &\textbf{62.76} &\textbf{46.95} &\textbf{63.84} &\textbf{45.77} \\
    BiGRU &56.83 &40.22 &61.84 &43.35 \\
    {RoBERTa$_{large}$} &60.30 &43.62 &60.97 &41.13 \\
    \hline
   
  \end{tabular}  }
  \label{tab:table4}
\end{table}

\subsubsection{Different Question Feature Encoding Models}
As indicated in Table~\ref{tab:table4}, we use different query encoders and the same pre-trained lanauage model RoBERTa$_{large}$ to test which query encoders is more effective. The experimental results in the table reveal that a simple LSTM model yields great performance.

    

\begin{table}[ht] 
  \caption{
    The effectiveness experiment of the QLSC method on Dev set and Over Sensitivity.
  }
  \centering
  \setlength{\tabcolsep}{3.8mm}
  \begin{tabular}{lcccc}
    \hline
      &\multicolumn{2}{c}{Dev set} &\multicolumn{2}{c}{Over Sensitivity}\\
     Model & $L1$ $\downarrow$& $L2$ $\downarrow$& $TCR$ $\uparrow$ & $TIR$ $\downarrow$ \\ 
    
    \hline
    RoBERTa &60.22 &2.42 &84.61 &8.55\\
    RoBERTa$_{+QLSC}$ &\textbf{12.28} &\textbf{0.54} &\textbf{86.72} &\textbf{3.44}\\
    \hline
   
  \end{tabular}  
  \label{tab:table_distance}
  
\end{table}

\subsubsection{Badcase Analysize}
Table~\ref{tab:table_distance} shows the effectiveness of the QLSC module in reducing the average distances of L1 and L2 for multiple paraphrased questions with the same semantics but different expressions. The average distances L1 and L2 are calculated by dividing the sum of the corresponding distances for each problem pair by the total number of samples.

A case study was conducted in the Dureader$_{robust}$ dataset, introducing two metrics, the Text Consistency Rate (TCR) and the Text Invalidity Rate (TIR), to evaluate the improvement in model robustness achieved by integrating the QLSC module. TCR measures the percentage of models that predict consistent answers for problems with different expressions but the same semantics. TIR gauges the proportion of empty text outputs among all answers.

The experiment revealed that the utilization of our QLSC module enhances TCR and reduces TIR. This substantiates that the inclusion of our auxiliary component effectively bolsters the robustness of the model. Following the implementation of our method, the model's extracted answers demonstrate improved consistency, whether they are accurate or erroneous, and also alleviate the problem of no-output answers.

\begin{table}[htbp]
\begin{CJK*}{UTF8}{gbsn}
 \caption{The badcase analysis on the Dureader$_{robust}$ dataset.}
  \centering
  \setlength{\tabcolsep}{0.3mm}
  \begin{tabular}{lcccccc}
    \hline
     question  && our && original && right \\ 
    \hline
    What is the constellation on November 16th? && Sagittarius &&Virgo && Scorpio\\
    What is the zodiac sign on November 16? && Sagittarius &&Libra && Scorpio\\
    \hline 
    Who is suitable to give the Gypsophila to? && lover && none && lover \\
    Who is Gypsophila suitable for? && lover && classmate && lover \\
    \hline
    
  \end{tabular}  
  \label{tab:table_case_analysis}

\end{CJK*}
\end{table}

Table~\ref{tab:table_case_analysis} presents a detailed analysis of the error cases. For the first example question, even though the model sometimes returned incorrect results, it provided consistent answers for various question formats. This consistency contributed to a significant reduction in the TIR value. The second example highlights that the QLSC can accurately answer variant-format questions, which the vanilla model finds unanswerable, thus decreasing the TIR.

\begin{figure}[t]
  \begin{center}
  \includegraphics[width=\linewidth]{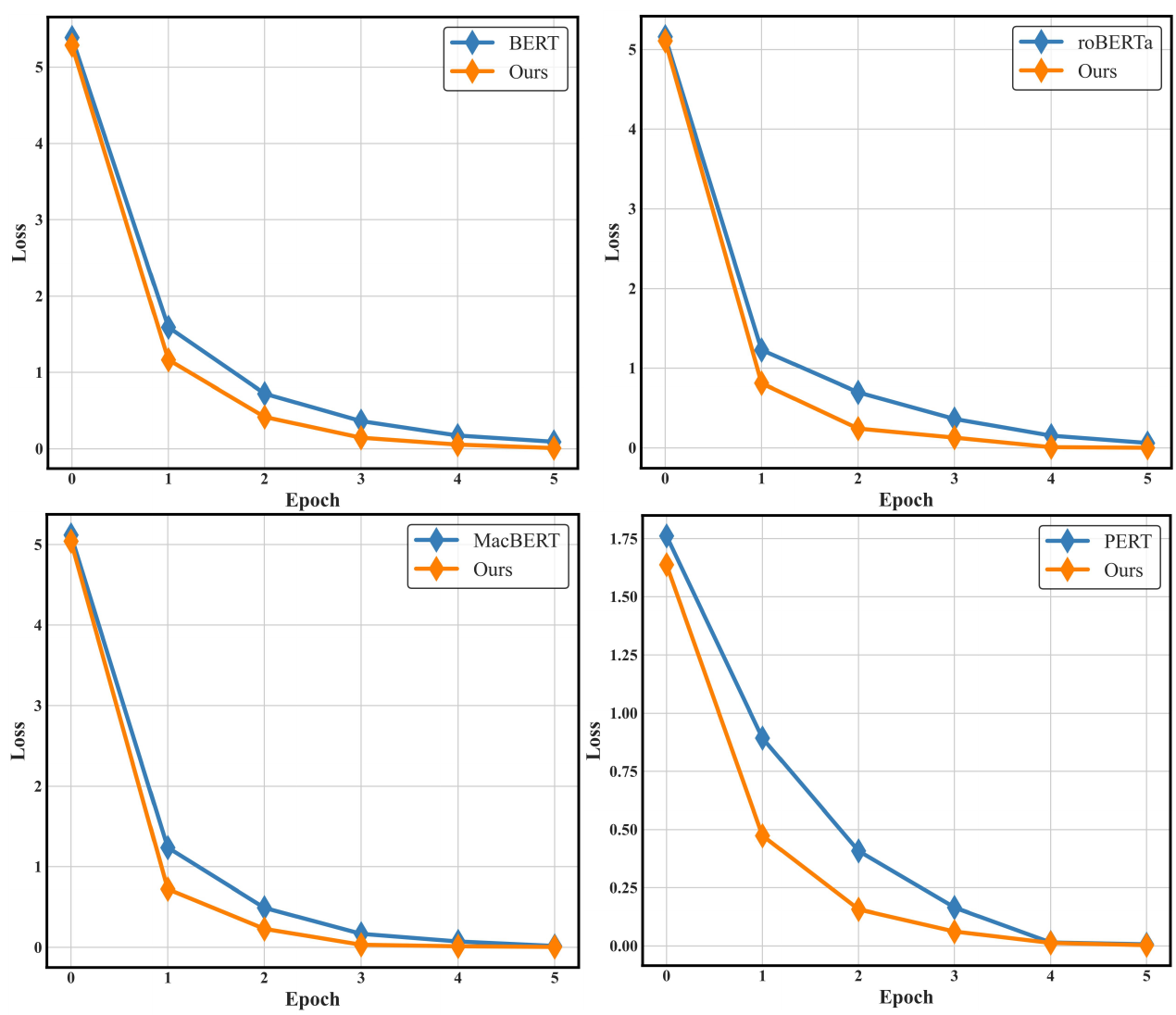}
  \end{center}
    \caption{The training loss of different models under each epochs.}
  \label{tab:table_loss}
\end{figure}

\begin{figure}[t]
  \begin{center}
  \includegraphics[width=\linewidth]{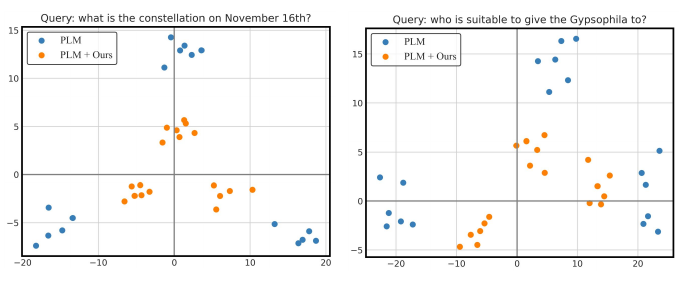}
  \end{center}
  \vspace{-1mm}
    \caption{PCA dimensionality reduction visualization results for querying embedded vectors.}
  \label{tab:table_query}
\end{figure}

\subsubsection{Visualized Experimental Analysis}
Figure~\ref{tab:table_loss} illustrates the results of training loss across different models and with the inclusion of the QLSC module as the number of epochs varies. The maximum number of epochs is limited to 5, and it is observed that the utilization of QLSC accelerates and improves model convergence during training.

Visualization analysis of the same set of rewritten query features was also performed after PCA dimensionality reduction. We performed a PCA experiment on two queries and their various format-variant queries, specifically the query "What is the constellation on November 16th?" and the query "Who is appropriate to present Gypsophila to?". In Figure~\ref{tab:table_query}, it can be seen that, with the use of the QLSC module, the distribution of the format-variant query features becomes closer and tighter.


\section{Conclusions}

In conclusion, our approach, the Query Latent Semantic Calibrator, effectively enhances the robustness of EQA models in handling format-variant inputs. By integrating latent semantic center features into the queries and passage embedding, our method improves the model's understanding of the queries-passage association. Extensive experiments show that our method accurately extracts answers from queries with different formats, but the same meaning. Our work highlights the importance of addressing robustness challenges in EQA and offers valuable insights for future research in improving machine reading comprehension.

\section{Acknowledgement}

This paper is supported by the Key Research and Development Program of Guangdong Province under grant No.2021B0101400003. Corresponding author is Xulong Zhang from Ping An Technology (Shenzhen) Co., Ltd. (zhangxulong@ieee.org).








\bibliographystyle{IEEEtran}
\bibliography{IEEEabrv,myrefs}

\end{document}